\documentclass[11pt]{article}

\usepackage[final]{acl}

\usepackage{times}
\usepackage{latexsym}

\usepackage[T1]{fontenc}
\usepackage{xcolor}

\usepackage[utf8]{inputenc}

\usepackage{microtype}

\usepackage{inconsolata}

\usepackage{graphicx}

\usepackage{caption}
\usepackage{subcaption}
\usepackage{booktabs}
\usepackage[table]{xcolor}
\usepackage{hyperref}

\usepackage{amsmath}

\usepackage{graphicx,amsmath,amssymb,booktabs,tabularx}
\usepackage{multirow}
\usepackage{hyperref}
\usepackage{xcolor}
\usepackage{subcaption}
\usepackage{tcolorbox}
\usepackage{makecell}
\usepackage{float}
\usepackage{calc}
\usepackage{pifont}
\usepackage{array}
\usepackage{multirow}
\usepackage{ragged2e}
\usepackage[table]{xcolor}
\usepackage{graphicx} 

\title{MARAG-R1: Beyond Single Retriever via Reinforcement-Learned Multi-Tool Agentic Retrieval}

\author{
Qi Luo\thanks{Equal contribution.} \quad
Xiaonan Li\footnotemark[1] \quad
Yuxin Wang  \quad
Tingshuo Fan \quad
Yuan Li \quad
Xinchi Chen \quad
Xipeng Qiu\thanks{Corresponding author.} \\
School of Computer Science, Fudan University \\
\texttt{qluo22@m.fudan.edu.cn} \quad \texttt{xpqiu@fudan.edu.cn}
}

\begin{document}
\maketitle

\begin{abstract}
Large Language Models (LLMs) excel at reasoning and generation but are inherently limited by static pretraining data, resulting in factual inaccuracies and weak adaptability to new information. Retrieval-Augmented Generation (RAG) addresses this issue by grounding LLMs in external knowledge; However, the effectiveness of RAG critically depends on whether the model can adequately access relevant information. Existing RAG systems rely on a single retriever with fixed top-$k$ selection, restricting access to a narrow and static subset of the corpus. As a result, this single-retriever paradigm has become the primary bottleneck for comprehensive external information acquisition, especially in tasks requiring corpus-level reasoning. To overcome this limitation, we propose \textbf{MARAG-R1}, a reinforcement-learned multi-tool RAG framework that enables LLMs to dynamically coordinate multiple retrieval mechanisms for broader and more precise information access. MARAG-R1 equips the model with four retrieval tools—semantic search, keyword search, filtering, and aggregation—and learns both \textit{how} and \textit{when} to use them through a two-stage training process: supervised fine-tuning followed by reinforcement learning. This design allows the model to interleave reasoning and retrieval, progressively gathering sufficient evidence for corpus-level synthesis. Experiments on GlobalQA, HotpotQA, and 2WikiMultiHopQA demonstrate that MARAG-R1 substantially outperforms strong baselines and achieves new state-of-the-art results in corpus-level reasoning tasks.
\end{abstract}


\section{Introduction}
Large Language Models (LLMs) have demonstrated remarkable reasoning and generative capabilities across diverse domains~
\cite{openai-o1,deepseek-r1}. However, their knowledge is limited to static pretraining data, which may lead to factual inaccuracies and reduced adaptability to new information~\cite{detectingllmhallucination, factuality_survey}. As a result, improving the ability to access and utilize external information has become a cornerstone for enhancing the factuality, interpretability, and trustworthiness of LLM-based reasoning.

\begin{figure}[t!]  
    \begin{flushright}  
        \captionsetup[subfigure]{justification=centering}
        \begin{subfigure}[b]{0.45\textwidth}
            \includegraphics[width=\textwidth]{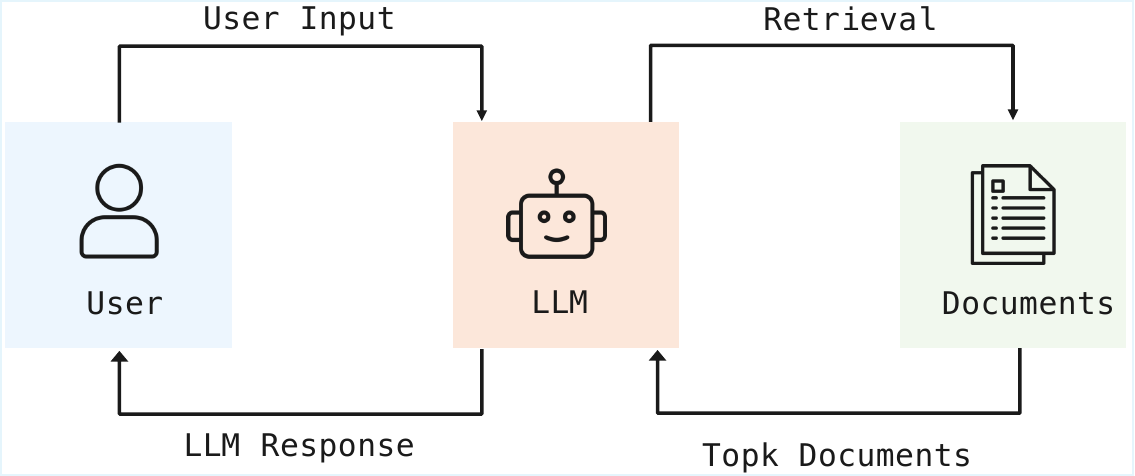}
            \caption{StandardRAG}
            \label{fig:MARAG_a}
        \end{subfigure}
        \hfill
        \begin{subfigure}[b]{0.45\textwidth}
            \includegraphics[width=\textwidth]{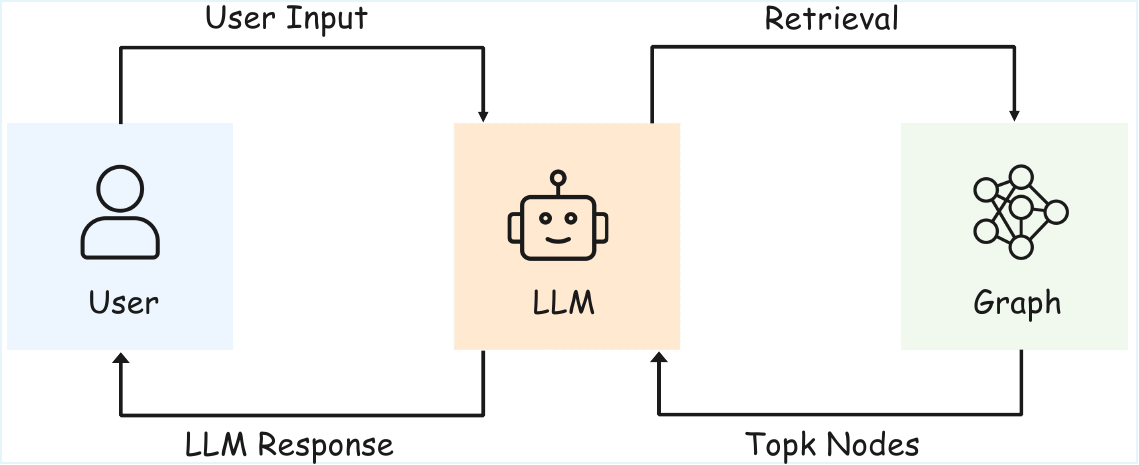}
            \caption{Graph-based RAG}
            \label{fig:MARAG_b}
        \end{subfigure}
        \hfill
        \begin{subfigure}[b]{0.45\textwidth}
            \includegraphics[width=\textwidth]{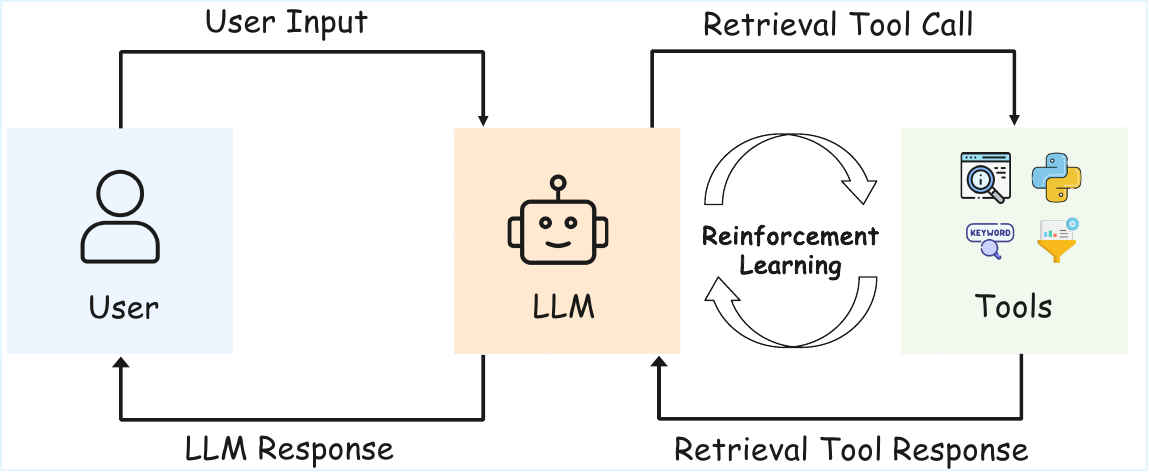}
            \caption{MARAG-R1}
            \label{fig:MARAG_c}
        \end{subfigure}
        \caption{
        Overview of the MARAG-R1 framework. 
        (a) The standard RAG model performs a single-round retrieval from a fixed top-$k$ document set, which often limits knowledge coverage. 
        (b) The graph-based RAG models structured semantic relations among documents, which enhances global awareness but loses the original document-level information during graph construction.
        (c) In contrast, our MARAG-R1 framework dynamically coordinates multiple specialized retrieval tools to access and integrate diverse external information, achieving more comprehensive and factual reasoning.
        }
        \label{fig:GRAG_overview}
    \end{flushright}
\end{figure}

A mainstream approach to addressing this limitation is Retrieval-Augmented Generation (RAG), which supplements LLMs with access to external knowledge sources during inference. By integrating retrieved evidence into the reasoning process, RAG mitigates hallucination and improves factual grounding.
Depending on how retrieval and reasoning are orchestrated, existing retrieval-augmented generation (RAG) frameworks can be broadly categorized into \textit{Workflow RAG} and \textit{Agentic RAG}. 
\textbf{Workflow RAG} systems (Figure~\ref{fig:MARAG_a}) adopt a fixed pipeline: given a query, the retriever selects a top-$k$ subset of documents, and the LLM generates an answer based solely on that static context~\cite{ircot,react,flare,self-rag}. 
While efficient, such workflows are inherently passive and limited in flexibility—once retrieval is complete, the model cannot refine or expand its evidence base. 
Recent evaluations show that even state-of-the-art retrievers achieve maximum recall below 0.45 on multi-hop QA tasks~\cite{hoprag}, and LLMs attain accuracy under 0.6 even when provided with gold evidence~\cite{multihoprag}. 
This rigidity fundamentally restricts the scalability of Workflow RAG in tasks that demand corpus-level reasoning and cross-document synthesis~\cite{graphrag,linearrag}. In contrast, \textbf{Agentic RAG} frameworks empower LLMs to actively plan, retrieve, and integrate information through iterative decision-making~\cite{search-r1,research,react}. 
Rather than following a predetermined pipeline, the model behaves as an autonomous agent that can issue queries, invoke tools, and refine its reasoning trajectory based on intermediate observations. 
Graph-based approaches (Figure~\ref{fig:MARAG_b}) such as GraphRAG~\cite{graphrag} and HyperGraphRAG~\cite{hypergraphrag} construct structured semantic abstractions to achieve global awareness, while reinforcement learning (RL)–based methods like Search-R1~\cite{search-r1} and ReSearch~\cite{research} optimize adaptive retrieval behaviors. 

However, existing RAG systems still suffer from \textit{limited external information access}.
Most approaches rely on a single retriever with a fixed top-$k$ selection strategy, which confines the model to a narrow and static subset of the corpus.
This restriction prevents the model from acquiring the full range of relevant external information required for tasks that demand comprehensive understanding across all documents, thereby making the current single-retriever paradigm a major bottleneck in external information acquisition.

To address this challenge, we introduce \textbf{MARAG-R1} (Figure~\ref{fig:MARAG_c}) — a unified framework for \textbf{M}ulti-tool \textbf{A}gentic \textbf{R}etrieval-\textbf{A}ugmented \textbf{G}eneration — that enables LLMs to acquire a wider range of external information through different retrieval tools and better utilize it via reinforcement learning to enhance global reasoning. Instead of relying on a single retriever or static graph abstraction, MARAG-R1 equips the model with four retrieval tools: \textit{semantic search} for relevance, \textit{keyword search} for precision, \textit{filtering} for constraint-based selection, and \textit{aggregation} for statistical synthesis. The model learns both \textit{when} and \textit{how} to use these tools through a two-stage training pipeline: (1) supervised fine-tuning provides an initial understanding of tool usage patterns, and (2) reinforcement learning optimizes tool sequencing and reasoning strategies for maximal coverage. This enables the model to interleave thinking and retrieval—progressively gathering and consolidating evidence until sufficient information is obtained—thereby bridging the gap between retrieval and global reasoning.

We summarize our contributions as follows:
\begin{itemize}
    \item To the best of our knowledge, MARAG-R1 is the first framework that enables an LLM to coordinate multiple retrieval tools through reinforcement learning to obtain more external information for global retrieval and reasoning.

     \item We conduct extensive comparisons with existing iterative and reinforcement-based RAG methods. Results show that MARAG-R1 significantly outperforms all baselines and establishes new state-of-the-art performance on global retrieval tasks.
    
    \item Further analyses reveal that each component contributes critically to overall improvements and exhibits strong transfer to multi-hop QA tasks.

\end{itemize}

\section{Related Work}
\subsection{Retrieval-Augmented Generation}

Retrieval-Augmented Generation (RAG) enhances language models by grounding their responses in external knowledge sources~\cite{rag,dpr}. Early RAG systems employed dense retrievers with dual-encoder architectures to retrieve relevant documents based on semantic similarity, achieving significant improvements in open-domain question answering~\cite{dpr,ms,nq,triviaqa}. Subsequent work focused on improving retrieval strategies through contrastive learning~\cite{contriever} and tighter integration of retrieval with generation~\cite{retro}. These foundational approaches primarily address single-hop retrieval tasks where answers reside in individual documents.

Recent advances have extended RAG to multi-hop reasoning scenarios requiring information synthesis across multiple documents. Self-RAG~\cite{self-rag} introduces self-reflection mechanisms that dynamically control retrieval timing and verify retrieved content quality. FLARE~\cite{flare} proposes active retrieval strategies that trigger document fetching based on generation confidence. IRCoT~\cite{ircot} interleaves chain-of-thought reasoning with iterative retrieval, demonstrating advantages on multi-hop reasoning benchmarks~\cite{musique,2wiki,hotpotqa}. While these methods improve reasoning over small document sets, they remain constrained by top-$k$ retrieval paradigms that fundamentally limit their ability to achieve complete document coverage for corpus-wide aggregation tasks~\cite{graphrag,hypergraphrag}.

Structured retrieval methods attempt to address limitations of unstructured approaches by organizing knowledge into graphs or hierarchical structures~\cite{graphrag,lightrag,hipporag}. KG-RAG~\cite{kg-rag} constructs knowledge graphs from unstructured text and performs multi-hop reasoning through graph traversal. GraphRAG~\cite{graphrag} builds multi-level community structures to support global queries requiring thematic synthesis across entire corpora. HippoRAG~\cite{hipporag} simulates hippocampal memory mechanisms using personalized PageRank for optimized retrieval paths. HyperGraphRAG~\cite{hypergraphrag} extends graph representations to hypergraphs that model complex multi-relational structures through hyperedges connecting multiple nodes. However, these graph-based approaches face fundamental challenges: information loss during graph construction limits their fidelity to original documents, and the locality of graph traversal restricts their ability to perform exhaustive corpus-wide retrieval required for global aggregation tasks~\cite{graphrag,hypergraphrag,kg-rag}.

\subsection{Agentic RAG Systems}

The emergence of tool-using agents has opened new possibilities for RAG systems. ReAct~\cite{react} demonstrates that language models can be prompted to interleave reasoning and action, enabling them to interact with external tools and APIs. Subsequent work has explored training agents specifically for retrieval tasks. LLatrieval~\cite{LaRetrieval} is a training-free retrieval framework where an LLM iteratively verifies and updates retrieval results through a verify–update process, enabling self-improvement without supervised training. These training-free agentic approaches show promise but are limited by the quality of prompting strategies and lack the ability to learn from task-specific feedback~\cite{flare,self-rag}. Recent work has begun exploring reinforcement learning~\cite{deepseek-r1, ppo, rloo, REINFORCE++} to optimize multi-step retrieval strategies. Search-R1~\cite{search-r1} and ReSearch~\cite{research} apply RL to train agents for iterative retrieval in multi-hop reasoning tasks, demonstrating improvements over fixed retrieval strategies. However, these methods optimize solely with answer-level rewards, providing sparse feedback that offers limited guidance for learning complex multi-step retrieval behaviors. More critically, they focus on optimizing local information retrieval within small document subsets rather than addressing the distinct challenge of exhaustive corpus-wide retrieval~\cite{graphrag,hypergraphrag}. Our work extends this line of research by introducing process-level rewards that provide dense supervision for intermediate retrieval steps and by explicitly targeting global retrieval tasks requiring complete document coverage and aggregation~\cite{search-r1,hipporag,r3-rag}.

\begin{figure*}[t]
    \centering
    \includegraphics[width=0.9\textwidth]{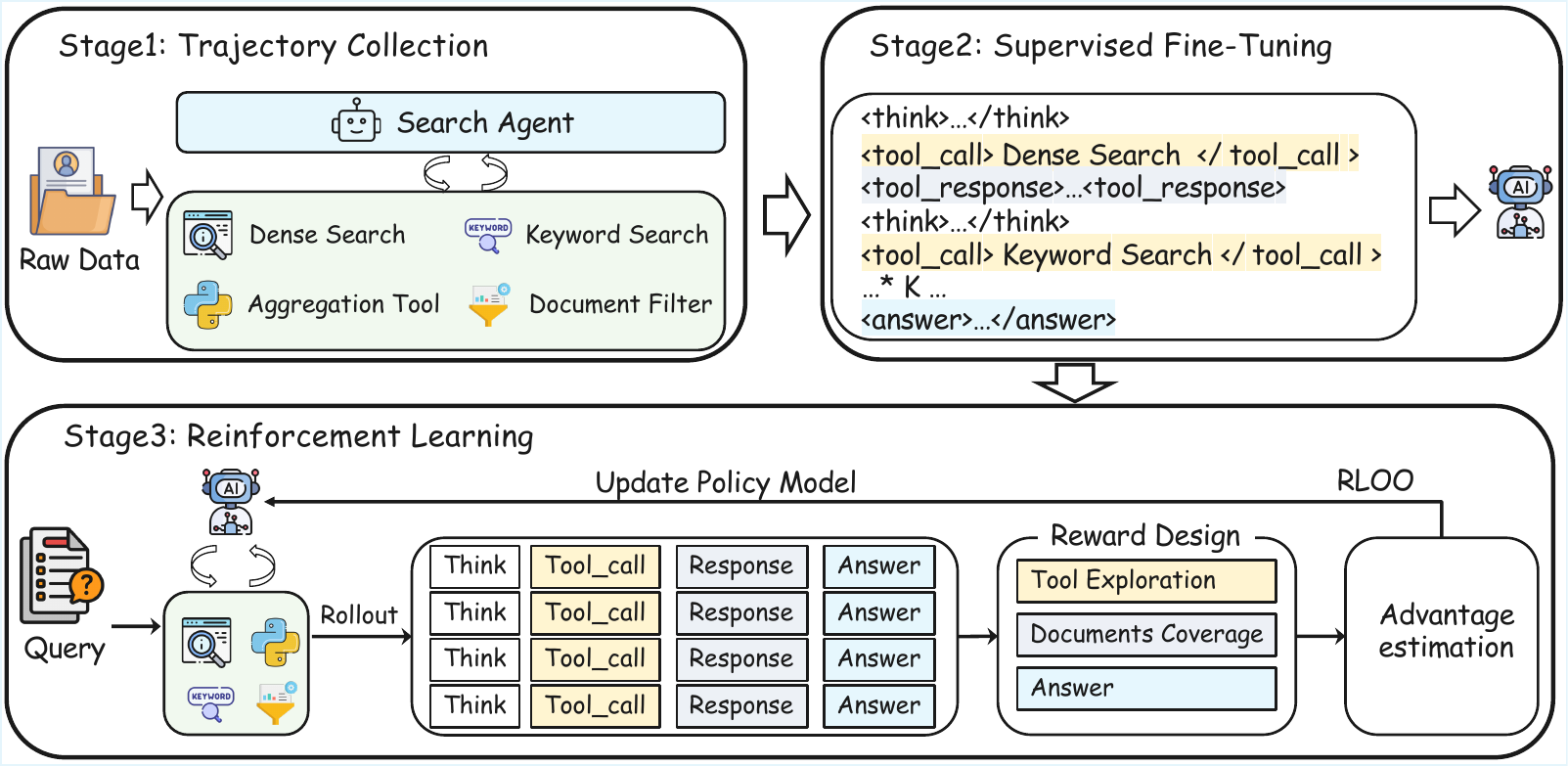}
    \captionsetup{justification=centering}
    \caption{Overview of the MARAG-R1 framework.}
    \label{fig:GRAG_overview}
\end{figure*}

\section{Method}
Instead of improving a single retriever, MARAG-R1 leverages multiple retrieval tools to acquire substantially more external information than a single retriever, and trains the LLM to better utilize this evidence for reasoning. As illustrated in Figure~\ref{fig:GRAG_overview}, MARAG-R1 consists of three major stages: (1) expert trajectory collection, (2) supervised fine-tuning, and (3) reinforcement learning for tool coordination. We next describe these three stages in detail.

\subsection{Stage 1: Trajectory Collection}

To train the agent effectively, we first collect high-quality expert trajectories that demonstrate multi-step reasoning for global RAG tasks. This stage involves two key components: designing specialized retrieval tools and defining the trajectory structure.

\subsubsection{Retrieval Tools}

To support diverse retrieval demands, MARAG-R1 is equipped with four complementary tools, each targeting a specific retrieval pattern:
\begin{enumerate}
    \item \textbf{Semantic Retriever $F_{\mathrm{DR}}$}: performs dense semantic retrieval for broad contextual exploration.
    \item \textbf{Keyword Retriever $F_{\mathrm{KR}}$}: performs keyword-based matching to retrieve relevant documents.
    \item \textbf{Document Filter $F_{\mathrm{DF}}$}: filters candidate documents based on metadata attributes or logical constraints.
    \item \textbf{Aggregation Tool $F_{\mathrm{AG}}$}: executes statistical or structural operations (e.g., counting, ranking, or set aggregation) to synthesize global evidence.
\end{enumerate}

\subsubsection{Trajectory Structure and Collection}

Each training instance consists of a query--answer pair $(Q, A^*)$. We expand it into a multi-step reasoning trajectory $\mathcal{T} = \{S_1, S_2, \ldots, S_{|\mathcal{T}|}\}$, where each step $S_t$ contains the model's intermediate reasoning $R_t$, the tool invocation $C_t$, and the resulting documents $D_t$:
\begin{equation}
S_t = (R_t, C_t, D_t), \quad t < |\mathcal{T}|.
\end{equation}
The final step replaces the tool call with the predicted answer $A$. Expert trajectories are collected using a teacher model under controlled prompting. Low-quality or redundant traces are filtered out through rejection sampling to guarantee trajectory consistency and factual correctness.

\subsection{Stage 2: Supervised Fine-Tuning}

We initialize the agent using expert trajectories generated by prompting GPT-4 with task instructions and examples, followed by rejection sampling for quality control. Given a dataset $\mathcal{D} = \{(Q_i, \mathcal{T}_i^*, A_i^*)\}_{i=1}^N$ of query-trajectory-answer triples, we optimize the standard next-token prediction loss:
\begin{equation}
\mathcal{L}_{\text{SFT}} = -\sum_{i=1}^{N} \sum_{j=1}^{|\mathcal{T}_i|} \log P_\theta(S_j \mid Q_i, S_{<j}),
\end{equation}
where $\theta$ represents the model parameters and $S_{<j}$ denotes all steps before step $j$. This stage teaches the agent to mimic expert behavior in tool selection and multi-step reasoning, establishing a strong foundation for subsequent RL training.
We refer to the model after this cold start stage as \emph{MARAG-CS} (SFT-only).

\subsection{Stage 3: Reinforcement Learning}
After imitation, MARAG-R1 is refined through reinforcement learning \cite{rloo} to optimize multi-tool coordination. The model interacts with the retrieval environment, autonomously generating trajectories and receiving rewards that reflect both outcome accuracy and procedural quality.
\subsubsection{Reward Design}

We design a composite reward that provides dense, interpretable feedback for both answer accuracy and retrieval behavior. 
Global retrieval requires not only producing correct final answers but also collecting comprehensive evidence through effective exploration of the document space. 
To this end, we decompose the total reward into three components: \emph{answer reward}, \emph{document coverage reward}, and \emph{tool exploration reward}.

\paragraph{Answer Reward ($R_A$).}
This term evaluates the correctness of the final answer using a token-level F1 score, which provides partial credit for answers that are semantically close to the reference:
\begin{equation}
R_A = \text{F1}(A, A^*),
\end{equation}
where $A$ and $A^*$ denote the predicted and ground-truth answers, respectively. 
This reward offers fine-grained supervision that encourages the agent to produce precise yet robust outputs.

\paragraph{Document Coverage Reward ($R_E$).}
To assess the completeness and precision of retrieved evidence, we compute an F1 score over document identifiers. 
Let $\mathcal{D}_{\text{pred}} = \{d_j\}_{j=1}^{|\mathcal{T}|-1}$ denote the set of retrieved document IDs and $\mathcal{D}^*$ the ground-truth supporting documents. 
The reward is defined as:
\begin{equation}
R_E = \text{F1}(\mathcal{D}_{\text{pred}}, \mathcal{D}^*) 
= \frac{2 \cdot P \cdot \text{Rec}}{P + \text{Rec}},
\end{equation}
where precision $P = \frac{|\mathcal{D}_{\text{pred}} \cap \mathcal{D}^*|}{|\mathcal{D}_{\text{pred}}|}$ 
and recall $\text{Rec} = \frac{|\mathcal{D}_{\text{pred}} \cap \mathcal{D}^*|}{|\mathcal{D}^*|}$. 
This reward encourages retrieving all necessary documents while minimizing irrelevant retrievals.

\paragraph{Tool Exploration Reward ($R_T$).}
This component promotes strategic exploration by rewarding sufficient but not excessive tool usage. 
Let $N_{\text{call}}$ and $N_{\text{call}}^*$ denote the number of tool calls in the predicted and expert trajectories, respectively:
\begin{equation}
R_T =
\begin{cases}
1, & \text{if } N_{\text{call}} \leq N_{\text{call}}^*, \\
\max\!\left(0,\, 1 - \frac{N_{\text{call}} - N_{\text{call}}^*}{N_{\text{call}}^*}\right), & \text{otherwise.}
\end{cases}
\end{equation}
This reward encourages the agent to explore the retrieval space beyond expert trajectories when beneficial, 
while penalizing redundant or unproductive tool calls to balance exploration with efficiency.

\paragraph{Overall Reward.}
The final trajectory-level reward aggregates the above components:
\begin{equation}
R(\mathcal{T}) = R_A + R_E + R_T.
\end{equation}

This hierarchical design prioritizes comprehensive evidence acquisition, 
which is essential for achieving global reasoning in RAG systems.



\subsubsection{Policy Optimization}

To optimize the policy, MARAG-R1 employs Reinforcement Learning with Leave-One-Out baseline (RLOO)~\cite{rloo}. 
Given a batch of $K$ trajectories $\{\mathcal{T}_i\}_{i=1}^{K}$ sampled from the current policy $\pi_\theta$, 
the gradient of the expected reward is estimated as:
\begin{equation}
\small
\begin{aligned}
\nabla_\theta J(\theta)
= \frac{1}{K} \sum_{i=1}^{K}
\Big(
R(\mathcal{T}_i)
- \tfrac{1}{K-1} \sum_{j \neq i} R(\mathcal{T}_j)
\Big)
\nabla_\theta \log P_\theta(\mathcal{T}_i).
\end{aligned}
\end{equation}

where $R(\mathcal{T}_i)$ denotes the scalar reward defined above. The leave-one-out baseline reduces gradient variance by comparing each trajectory’s reward to the mean of others within the same batch, 
thus stabilizing optimization without requiring an explicit reference model or KL regularization. This formulation is particularly suitable for multi-step reasoning tasks with discrete tool-use actions, 
as it allows efficient credit assignment at the trajectory level. Through RLOO optimization, MARAG-R1 refines its retrieval and reasoning strategies beyond supervised imitation, 
achieving more consistent and efficient multi-tool coordination.

\section{Experimental Setup}

\subsection{Baselines}
We compare our approach with six representative Retrieval-Augmented Generation (RAG) methods~\cite{rag,ircot,graphrag,hypergraphrag,search-r1,research}. 
StandardRAG~\cite{rag,dpr} represents the basic single-round retrieval paradigm. 
ITER-RETGEN~\cite{iter-retgen} adopts an iterative retrieval–generation framework that alternates between content generation and evidence retrieval. 
IRCoT~\cite{ircot} performs multi-round retrieval refinement through iterative question decomposition. 
HyperGraphRAG~\cite{hypergraphrag} enhances graph-based reasoning by optimizing the construction of knowledge graphs in GraphRAG~\cite{graphrag}. 
Search-R1~\cite{search-r1} applies reinforcement learning to optimize graph-based retrieval decisions. 
ReCall~\cite{research} leverages reinforcement learning to coordinate the same multi-tool retrieval suite as ours; unlike MARAG-R1, it is optimized solely with answer-level rewards.

\subsection{Datasets}
To comprehensively evaluate the proposed MARAG-R1 framework across different reasoning granularities, we employ both corpus-level and multi-hop benchmarks. 
Our main experiments are conducted on the GlobalQA benchmark~\cite{graphrag}, which focuses on corpus-level reasoning and information aggregation. 
GlobalQA consists of four task types—\textbf{TopK}, \textbf{Count}, \textbf{Sort}, and \textbf{MinMax}—that require global reasoning and information integration across the entire corpus~\cite{graphrag,hypergraphrag}. To further analyze model generalization and the effect of different components, we perform ablation studies on two widely used multi-hop reasoning datasets: 
2WikiMultiHopQA~\cite{2wiki}, which requires reasoning across multiple Wikipedia documents, and 
HotpotQA~\cite{hotpotqa}, which focuses on identifying supporting evidence from multiple sources~\cite{ircot,flare}. 
This dual evaluation setup allows us to verify that MARAG-R1 not only excels in corpus-level aggregation but also maintains strong reasoning ability in traditional multi-hop QA scenarios.

\subsection{Evaluation Metrics}
We adopt two metrics to evaluate performance:  

\textbf{F1}: measures the quality of final answers, using the standard token-level F1 score:  
\begin{equation}
\text{F1} = 2 \cdot \frac{\text{Precision} \cdot \text{Recall}}{\text{Precision} + \text{Recall}}
\end{equation}

\textbf{Document F1@k(D-F1@k)}: evaluates the coverage of retrieved documents, defined as the F1 score between the retrieved document set and the gold document set:  
\begin{equation}
\text{Precision@k} = \frac{|\mathcal{D}\text{ret}^k \cap \mathcal{D}\text{gold}|}{|\mathcal{D}_\text{ret}^k|}
\end{equation}\begin{equation}
\text{Recall@k} = \frac{|\mathcal{D}\text{ret}^k \cap \mathcal{D}\text{gold}|}{|\mathcal{D}_\text{gold}|}
\end{equation}\begin{equation}
\text{D-F1@k} = \frac{2 \cdot \text{Precision@k} \cdot \text{Recall@k}}{\text{Precision@k} + \text{Recall@k}}
\end{equation}

\subsection{Implementation Details}
We unify the experimental configuration across all methods to ensure fair comparison. We employ BGE (BAAI General Embedding)~\cite{bge} as the retriever and Qwen3-4B~\cite{qwen3} as the filtering model, with TopK set to 20 for retrieval. 
All baseline and proposed models are executed under identical hardware and hyperparameter settings~\cite{rag,search-r1,research}. We adopt the chunk-free retrieval strategy~\cite{chunkfree} to maintain consistent evidence granularity across systems. For reinforcement learning, we follow the RLOO optimization scheme with 5 sampled rollouts per query. Evaluation follows standard metrics: F1 for token-level answer accuracy and D-F1@20, which measures document-level evidence coverage at k=20.

\begin{table*}[t]
\centering
\small
\resizebox{1.0\textwidth}{!}{%

\begin{tabular}{lcrrrrrrrrrr}

\toprule
Method & Infer Env & \multicolumn{2}{c}{TopK} & \multicolumn{2}{c}{Count} & \multicolumn{2}{c}{Sort} & \multicolumn{2}{c}{MinMax} & \multicolumn{2}{c}{Avg} \\
\cmidrule(lr){3-9}  \cmidrule(lr){9-10} \cmidrule(lr){11-12}
 & & F1 & D-F1@20 & F1 & D-F1@20 & F1 & D-F1@20 & F1 & D-F1@20 & F1 & D-F1@20 \\
 \midrule
 \rowcolor{gray!15}
\multicolumn{12}{c}{Qwen2.5-3B-Instruct} \\
\includegraphics[height=0.75em]{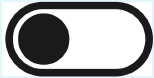} StandardRAG & \raisebox{-0.2\height}{\includegraphics[height=1em]{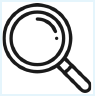}} & 2.14 & 7.98 & 0.93 & 8.88 & 2.58 & 7.65 & 0.80 & 7.99 & 1.56 & 8.09 \\
\includegraphics[height=0.75em]{pictures/off.pdf}IRCoT &\raisebox{-0.2\height}{\includegraphics[height=1em]{pictures/search-icon.pdf}} & 1.67 & 10.01 & 1.09 & 10.3 & 1.56 & 11.03 & 1.60 & 10.44 & 1.52 & 10.38 \\
\includegraphics[height=0.75em]{pictures/off.pdf}HyperGraphRAG &\raisebox{-0.2\height}{\includegraphics[height=1em]{pictures/search-icon.pdf}} & 0.06 & - &0.39 &  - & 0.04 & -   & 0.00 & - &  0.09 & -   \\

\includegraphics[height=0.75em]{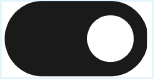} Search-R1 &\raisebox{-0.2\height}{\includegraphics[height=1em]{pictures/search-icon.pdf}} &1.29	&10.02	&1.40	&11.84&	1.95&	11.06&	2.40&	8.56	&1.78	&10.07  \\
\includegraphics[height=0.75em]{pictures/on.pdf} ReCall &\raisebox{-0.2\height}{\includegraphics[height=1em]{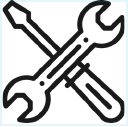}} & 4.52 &  6.74 & 1.40 & 16.81  & 13.62 &  15.55 & 7.20 & 14.54  & 6.41 & 12.68 \\
\includegraphics[height=0.75em]{pictures/off.pdf} MARAG & \raisebox{-0.2\height}{\includegraphics[height=1em]{pictures/tool-icon.pdf}} & 1.64 & 3.76 & 0.93 & 14.61 & 9.19 & 14.37 & 4.18 & 7.52 & 3.66 & 8.85 \\
\includegraphics[height=0.75em]{pictures/on.pdf} MARAG-CS & \raisebox{-0.2\height}{\includegraphics[height=1em]{pictures/tool-icon.pdf}} & 27.23 & 34.82 & 5.58 & 37.92 & 35.61 & 37.12 & 32.27 & 37.26 & 26.32 & 36.59 \\
\includegraphics[height=0.75em]{pictures/on.pdf} \textbf{MARAG-R1} & \raisebox{-0.2\height}{\includegraphics[height=1em]{pictures/tool-icon.pdf}} & \textbf{27.82} & \textbf{34.24} & \textbf{5.58} & \textbf{38.52} & \textbf{33.03} & \textbf{35.15} & \textbf{33.33} & \textbf{39.98} & \textbf{26.40} & \textbf{37.05} \\
\midrule
\rowcolor{gray!15}
\multicolumn{12}{c}{Qwen2.5-7B-Instruct} \\

\includegraphics[height=0.75em]{pictures/off.pdf} StandardRAG &\raisebox{-0.2\height}{\includegraphics[height=1em]{pictures/search-icon.pdf}} & 1.64 & 7.98   & 0.00 & 8.88 & 2.73 & 7.65 & 1.33 & 7.99 & 1.43 & 8.09 \\
\includegraphics[height=0.75em]{pictures/off.pdf} IRCoT &\raisebox{-0.2\height}{\includegraphics[height=1em]{pictures/search-icon.pdf}} & 1.30 & 7.76 & 0.04 & 8.68 & 0.96 & 7.68 & 1.34 & 8.25 & 1.02 & 8.07 \\
\includegraphics[height=0.75em]{pictures/off.pdf} HyperGraphRAG &\raisebox{-0.2\height}{\includegraphics[height=1em]{pictures/search-icon.pdf}} & 0.40 & - & 0.46 & - & 0.23 & -   & 0.00 & - & 0.25 & - \\
\includegraphics[height=0.75em]{pictures/on.pdf} Search-R1 &\raisebox{-0.2\height}{\includegraphics[height=1em]{pictures/search-icon.pdf}} &4.74	&8.46&	0.93&	9.20	&0.33&	10.01&	3.73&	9.40	&2.93&	9.17\\
\includegraphics[height=0.75em]{pictures/on.pdf} ReCall &\raisebox{-0.2\height}{\includegraphics[height=1em]{pictures/tool-icon.pdf}} & 12.07 & 14.53 & 2.33  & 23.63  &22.84  & 22.90  & 8.00 & 14.29 & 10.84 & 17.61 \\

\includegraphics[height=0.75em]{pictures/off.pdf} MARAG & \raisebox{-0.2\height}{\includegraphics[height=1em]{pictures/tool-icon.pdf}} & 3.55 & 6.52 & 1.40 & 12.79 & 13.00 & 14.85 & 2.67 & 3.58 & 4.53 & 8.20 \\
\includegraphics[height=0.75em]{pictures/on.pdf} MARAG-CS & \raisebox{-0.2\height}{\includegraphics[height=1em]{pictures/tool-icon.pdf}} & 27.55 & 34.12 & 6.05 & 39.13 & 37.95 & 38.41 & 33.60 & 41.50 & 27.35 & 38.19 \\
\includegraphics[height=0.75em]{pictures/on.pdf} \textbf{MARAG-R1} & \raisebox{-0.2\height}{\includegraphics[height=1em]{pictures/tool-icon.pdf}} & \textbf{30.24} & \textbf{36.15} & \textbf{7.44} & \textbf{40.44} & \textbf{38.28} & \textbf{38.06} & \textbf{33.87} & \textbf{39.72} & \textbf{28.60} & \textbf{38.44} \\
\rowcolor{gray!15}
\midrule
\multicolumn{12}{c}{Qwen2.5-14B-Instruct} \\
\includegraphics[height=0.75em]{pictures/off.pdf} StandardRAG &\raisebox{-0.2\height}{\includegraphics[height=1em]{pictures/search-icon.pdf}} & 1.78 & 7.98 & 0.93 & 8.88 & 1.96 & 7.65 & 1.33 & 7.99 & 1.51 & 8.09 \\
\includegraphics[height=0.75em]{pictures/off.pdf} IRCoT &\raisebox{-0.2\height}{\includegraphics[height=1em]{pictures/search-icon.pdf}} & 0.13 & 8.86 & 0.01 & 8.52 & 0.25 & 9.56   & 0.00 & 8.38 & 0.09 & 8.77 \\
\includegraphics[height=0.75em]{pictures/off.pdf} HyperGraphRAG &\raisebox{-0.2\height}{\includegraphics[height=1em]{pictures/search-icon.pdf}} & 0.00  & - & 0.47  & - & 0.04 & -   & 0.00 & - & 0.09 & - \\

\includegraphics[height=0.75em]{pictures/on.pdf} Search-R1 &\raisebox{-0.2\height}{\includegraphics[height=1em]{pictures/search-icon.pdf}} &4.74&	8.46&	0.93	&9.20	&0.33&	10.01&	3.73	&9.40&	2.93	&9.17  \\
\includegraphics[height=0.75em]{pictures/on.pdf} ReCall &\raisebox{-0.2\height}{\includegraphics[height=1em]{pictures/tool-icon.pdf}} & 15.07 & 20.52 & 2.79 & 19.67 & 23.59 & 23.21 & 14.93 & 17.92 & 14.25 & 20.00  \\
\includegraphics[height=0.75em]{pictures/off.pdf} MARAG & \raisebox{-0.2\height}{\includegraphics[height=1em]{pictures/tool-icon.pdf}} & 8.62 & 14.06 & 2.79 & 15.19 & 16.45 & 16.23 & 7.73 & 10.73 & 8.63 & 13.58 \\
\includegraphics[height=0.75em]{pictures/on.pdf} MARAG-CS & \raisebox{-0.2\height}{\includegraphics[height=1em]{pictures/tool-icon.pdf}} & 31.60 & 37.62 & 6.05 & 40.37 & 38.22 & 38.39 & 36.53 & 42.45 & 28.92 & 39.83 \\
\includegraphics[height=0.75em]{pictures/on.pdf} \textbf{MARAG-R1} & \raisebox{-0.2\height}{\includegraphics[height=1em]{pictures/tool-icon.pdf}} & \textbf{32.65} & \textbf{38.81} & \textbf{6.05} & \textbf{42.38} & \textbf{40.5} & \textbf{40.92} & \textbf{39.20} & \textbf{45.81} & \textbf{31.22} & \textbf{42.11} \\
\bottomrule

\end{tabular}%
}

\caption{
Main results on the \textbf{GlobalQA} benchmark using three model scales of Qwen2.5 (3B, 7B, and 14B). 
The GlobalQA benchmark comprises four task types: \textbf{TopK}, \textbf{Count}, \textbf{Sort}, and \textbf{MinMax}, 
which require global reasoning and information aggregation across entire corpus. 
\protect\includegraphics[height=0.75em]{pictures/on.pdf} indicates trained models, while 
\protect\includegraphics[height=0.75em]{pictures/off.pdf} denotes non-trained variants. 
\protect\includegraphics[height=1em]{pictures/search-icon.pdf} represents inference environments with a single retriever, 
and \protect\includegraphics[height=1em]{pictures/tool-icon.pdf} indicates multi-tool retrieval environments. 
F1 measures answer accuracy, and D-F1@k measures execution accuracy over retrieved documents.
}

\label{tab:main_results}
\end{table*}

\section{Main Results}

We present the main results in Table~\ref{tab:main_results}. 
Experiments are conducted on four tasks in \textbf{GlobalQA} using three scales of Qwen2.5 models (3B, 7B, and 14B parameters) to assess performance across varying model capacities. 
F1 measures final answer accuracy, while D-F1@k (Execution F1) evaluates the correctness of intermediate reasoning steps.

\paragraph{Overall Performance.}
MARAG-R1 achieves state-of-the-art results across all model scales, outperforming all existing retrieval and reasoning frameworks. Across Qwen2.5-3B, 7B, and 14B, our model consistently attains the highest F1 scores (26.4 → 28.6 → 31.22), indicating superior end-task accuracy. More importantly, MARAG-R1 also achieves the best D-F1@20 results (37.05 → 38.44 → 42.11), which measure the correctness of intermediate reasoning steps and thereby reflect the model’s ability to acquire and utilize external information. By leveraging our reinforcement-learned multi-tool coordination structure, MARAG-R1 is able to gather and integrate substantially more external information, thereby achieving more accurate and consistent reasoning outcomes.

\paragraph{Comparison with Graph-based Methods.}

Compared with HyperGraphRAG, MARAG-R1 exhibits clear superiority across all tasks.
While graph-based methods aim to approximate global reasoning through hierarchical graph structures or summary nodes, such abstractions inevitably lose fine-grained factual and numerical details.
As a result, HyperGraphRAG fails to produce meaningful F1 scores on most tasks, whereas MARAG-R1 achieves 32.65 on \textit{TopK} and 39.2 on \textit{MinMax}.
This highlights that constructing static graphs cannot ensure full information coverage or compositional reasoning accuracy.
In contrast, MARAG-R1 performs dynamic retrieval and synthesis, using explicit tool calls to access and aggregate dispersed evidence at runtime.

\paragraph{Comparison with Local RAG Methods.}

Local RAG methods—including StandardRAG, IRCoT, and Search-R1—can interleave reasoning and retrieval, but remain constrained by local retrieval scopes.
Their retrievers typically operate on a small set of top-ranked documents per query and use only a single retriever as the external knowledge interface.This narrow access scope restricts exposure to the full corpus and hinders comprehensive aggregation or statistical synthesis across dispersed evidence. In contrast, MARAG-R1 equips the model with multiple complementary retrieval tools—semantic search, keyword search, filtering, and aggregation—enabling it to iteratively gather and combine information from diverse perspectives. By dynamically coordinating these tools, MARAG-R1 expands the effective retrieval horizon beyond local contexts and supports corpus-level reasoning that traditional single-retriever frameworks cannot achieve.

\paragraph{Effect of Reinforcement Learning.}
The improvement of MARAG-R1 over its supervised-only counterpart (\textbf{MARAG-CS}) verifies the effectiveness of our two-stage training strategy. 
In Qwen2.5-14B, MARAG-R1 improves the average F1 from 28.92 to 31.22 and D-F1@k from 39.83 to 42.11. 
Gains are particularly notable on the \textit{MinMax} task (39.2 / 45.81 vs.\ 36.53 / 42.45), indicating that our process-level reward facilitates deeper multi-step reasoning. 
Smaller yet consistent improvements are also observed on \textit{Sort} and \textit{TopK}, confirming that decomposed rewards provide finer guidance beyond final-answer supervision.

\paragraph{Scaling Behavior}
All methods benefit from larger model sizes, yet MARAG-R1 not only scales better but also consistently achieves state-of-the-art performance across all capacities.
From 3B to 14B, its F1 improves from 26.4 to 31.22, while \textbf{ReCall}—despite operating under the same tool-augmented environment—only rises modestly from 6.41 to 14.25.
The results suggest that access to multiple tools alone is not sufficient, and that learning appropriate strategies for tool utilization is necessary.
The superior scaling of MARAG-R1 highlights the effectiveness of our two-stage training pipeline and reward design, which jointly enable the model to progressively refine retrieval, reasoning, and aggregation behaviors as capacity increases.

\begin{table}[t]
\begin{minipage}[t]{0.98\linewidth}
\centering
\renewcommand{\arraystretch}{1.2}
\setlength{\tabcolsep}{5pt}
\begin{tabular}{lcc}
\hline

\textbf{Configuration} & \textbf{F1} & \textbf{D-F1@20} \\
\hline
MARAG-R1 (Full) & \textbf{31.22} & \textbf{42.11} \\
\quad w/o Tool Calls Reward & 29.45 & 39.54 \\
\quad w/o Document Reward & 28.87 & 39.53 \\
\quad w/o Answer Reward & 13.78 & 24.48 \\
\quad w/o SFT & 14.25 & 20.00 \\

\hline
\end{tabular}
\caption{Ablation study results. ``w/o'' denotes removal of the corresponding component.}
\label{tab:ablation_componet}
\end{minipage}
\end{table}

\section{Analysis}
\subsection{Impact of Training Components}
To assess the contribution of each component in MARAG-R1, we conduct ablation experiments by removing individual reward terms and the supervised fine-tuning (SFT) stage.
Table~\ref{tab:ablation_componet} reports results on Qwen2.5-14B across GlobalQA tasks. Removing the SFT stage leads to the largest drop of 17.0 F1 and 22.1 D-F1@20, showing that SFT provides essential initialization for stable policy learning and effective tool-use exploration.
Eliminating the answer reward causes a similar decrease of 17.4 F1 and 17.6 D-F1@20, indicating that end-task supervision remains the key signal for aligning reasoning with correct answers. Removing the document reward results in a drop of 2.4 F1 and 2.6 D-F1@20, demonstrating its importance in ensuring full evidence coverage and preventing shallow top-k retrieval. Omitting the tool-calls reward slightly reduces performance by 1.8 F1 and 2.6 D-F1@20, confirming its role in guiding more tool use to get external information. 
Overall, these ablations show that every component contributes complementary supervision: SFT stabilizes training, the answer reward drives correctness, and the document and tool-calls rewards enhance retrieval completeness and efficiency—jointly enabling MARAG-R1’s strong global reasoning performance.

\subsection{Impact of External Information Acquisition}
\begin{table}[t]
\begin{minipage}[t]{0.98\linewidth}
\centering
\renewcommand{\arraystretch}{1.2}
\setlength{\tabcolsep}{5pt}
\begin{tabular}{lrrr}
\hline
\textbf{Method} & \textbf{F1} & \textbf{D-F1@20} & \textbf{Avg. Calls} \\
\hline
ReCall & 14.25 & 20.00 & 4.53 \\
MARAG & 8.63 & 13.58 & 2.81 \\
MARAG-CS & 28.92 & 39.83 & 6.27 \\
MARAG-R1 & \textbf{31.22} & \textbf{42.11} & \textbf{6.32} \\
\hline
\end{tabular}
\caption{Performance and average number of tool calls on Qwen2.5-14B.}
\label{tab:tool_calls}
\end{minipage}
\end{table}

In RAG, model performance fundamentally depends on how much and how effectively external information is acquired during reasoning. The goal of our two-stage training process (SFT + RL) is precisely to enhance this ability. Through reward design and policy optimization, the model learns to proactively and persistently invoke retrieval tools to gather additional evidence that supports intermediate and final reasoning steps. As shown in Table~\ref{tab:tool_calls}, MARAG-R1 acquires substantially more external information while maintaining coherent reasoning. It achieves the highest average number of tool calls (6.32 per query) and correspondingly the best F1 (31.22) and D-F1@20 (42.11) scores. In contrast, MARAG performs far fewer calls (2.81), resulting in much lower performance (8.63 / 13.58), suggesting that insufficient retrieval severely limits information completeness. Although MARAG-CS makes a similar number of calls (6.27), MARAG-R1 still surpasses it by +1.6 F1 and +2.3 D-F1@20, indicating that the reinforcement learning stage not only increases call frequency but also improves call quality and sequencing. The model learns when and how to issue additional calls to enrich evidence, retrieve complementary documents, and integrate them effectively into subsequent reasoning. In summary, the improvement stems from a learned ability to acquire and exploit more external information. MARAG-R1’s training teaches the model to make purposeful tool invocations that expand the available evidence base—ultimately driving stronger and more informed reasoning.
\subsection{Impact of Different Retrieval Tools}
To assess the contribution of each retrieval tool, we perform an ablation study by systematically removing individual tools from MARAG-R1.
Following the same two-stage training protocol (SFT + RL), each variant excludes one of the four core tools—Document Filter, Semantic Search, Keyword Retrieval, or Aggregation Function—and is evaluated on the GlobalQA benchmark using Qwen2.5-7B-Instruct. During inference, the model can only coordinate the remaining tools to complete the task. Results in Table~\ref{tab:tool-ablation} show a clear hierarchy of importance among the tools.
Removing the Aggregation Function leads to the largest performance drop (↓16.9 F1, ↓18.7 D-F1@20), as the model loses the ability to perform corpus-level statistical reasoning essential for Count, Sort, and MinMax tasks. Eliminating Keyword Retrieval results in a decline of 14.3 F1 and 14.1 D-F1@20, reflecting its critical role in precise entity or attribute matching (e.g., filtering by organization or location).
Removing Semantic Search causes a 7.1 F1 and 13.1 D-F1@20 drop, indicating its importance for semantic coverage and natural-language-based document retrieval. Even the Document Filter, which governs metadata-based constraints, contributes noticeably (-4.8 F1 and -6.0 D-F1@20), helping narrow the search space for efficient reasoning. Overall, these results demonstrate that no single tool is sufficient for effective global reasoning —— Different tools provide complementary external information, enabling MARAG-R1 to effectively fectively accomplish
global retrieval tasks.

\begin{table}[t]
\begin{minipage}[t]{\linewidth}
\centering
\setlength{\tabcolsep}{4pt}
\begin{tabular}{lcc}
\hline
\textbf{Configuration} & \textbf{F1} & \textbf{D-F1@20} \\
\hline
MARAG-R1 (Full) & \textbf{27.35} & \textbf{38.19} \\
\quad w/o Document Filter & 22.57 & 32.14 \\
\quad w/o Semantic Retriever & 20.21 & 25.13 \\
\quad w/o Keyword Retriever & 13.03 & 24.08 \\
\quad w/o Aggregation Function & 10.48 & 19.54 \\
\hline
\end{tabular}
\caption{Ablation study on different retrieval tools in MARAG-R1 using Qwen2.5-7B-Instruct.}
\label{tab:tool-ablation}
\end{minipage}
\end{table}

\subsection{Generalization to Multi-Hop Reasoning Tasks}
\begin{table}[t]
\begin{minipage}[t]{1\linewidth}
\centering
\small
\renewcommand{\arraystretch}{1.3}
\setlength{\tabcolsep}{5pt}
\resizebox{\linewidth}{!}{
\begin{tabular}{l*{6}{c}}
\hline
\multirow{2}{*}{\textbf{Method}} & \multicolumn{2}{c}{\textbf{2Wiki}} & \multicolumn{2}{c}{\textbf{HotpotQA}} & \multicolumn{2}{c}{\textbf{Avg.}} \\
\cline{2-5} \cline{6-7}
 & \textbf{EM} & \textbf{F1} & \textbf{EM} & \textbf{F1} & \textbf{EM} & \textbf{F1} \\
\hline
StandRAG     & 12.00 & 18.24 & 19.00 & 26.36 & 15.50 & 22.30 \\
ITER-RETGEN   & 12.00 & 17.22 & 17.00 & 25.56 & 14.50 & 21.39 \\
IRCoT        & 5.00  & 18.50 & 11.00 & 23.92 & 8.00  & 21.21 \\
\hline
MARAG    & 15.00 & 22.81 & 22.00 & 34.05 & 18.50 & 28.43 \\
MARAG-R1 & \textbf{22.00} & \textbf{26.93} & \textbf{31.00} & \textbf{39.16} & \textbf{26.50} & \textbf{33.05} \\
\hline
\end{tabular}
}
\caption{Performance comparison on multi-hop question answering datasets}
\label{tab:multihop}
\end{minipage}
\end{table}

To evaluate the generalization ability of MARAG-R1, we test it on two unseen multi-hop QA datasets: 2WikiMultiHopQA and HotpotQA.
Although these benchmarks primarily target local retrieval, they still require multi-step reasoning across multiple documents, making them suitable for assessing transfer from global retrieval training. As shown in Table~\ref{tab:multihop}, MARAG-R1 substantially outperforms all baselines on both datasets. It achieves 22.00 EM / 26.93 F1 on 2WikiMultiHopQA and 31.00 EM / 39.16 F1 on HotpotQA, surpassing the strongest baseline (IRCoT) by large margins across all metrics. Averaged over datasets, MARAG-R1 improves EM by +18.5 and F1 by +11.8, confirming strong cross-task transfer. These gains highlight that MARAG-R1’s multi-tool coordination and adaptive retrieval strategies—learned from global aggregation tasks—generalize effectively to conventional multi-hop reasoning.
Unlike iterative baselines that follow fixed retrieval patterns, MARAG-R1 dynamically balances exploration and precision, selecting the most relevant tools based on intermediate context. Overall, the model demonstrates robust generalization: the retrieval reasoning skills acquired through reinforcement learning are not tied to specific datasets, but transferable to broader multi-step reasoning tasks.

\subsection{Case Study}
To illustrate the qualitative difference between graph-based and tool-based reasoning, we present a case study comparing HyperGraphRAG and MARAG-R1 on an OR-type query (Table~\ref{tab:hypergragrag_case_study}). In HyperGraphRAG, the model retrieves the top-10 most relevant nodes and edges during the first step, achieving high coherence scores ($\approx$1.0).
However, the predefined graph structure causes severe information loss—document-level boundaries and identifiers are no longer preserved.
As a result, the system cannot determine which original documents the retrieved snippets belong to and fails to perform the required aggregation, ultimately producing an incorrect answer (ID = 1).
This highlights a fundamental limitation of static graph representations: they compress the corpus into symbolic abstractions at the cost of fine-grained evidence traceability. In contrast, MARAG-R1 executes a transparent sequence of tool calls that directly operate on the underlying corpus. The model performs separate searches for each subquery, unifies their results through explicit set operations, and finally applies an aggregation function to compute the minimal document ID.
This dynamic reasoning process preserves both the granularity and provenance of retrieved evidence, leading to the correct answer (ID = 21). Overall, this case demonstrates how MARAG-R1’s explicit tool orchestration enables faithful and interpretable reasoning, effectively overcoming the information bottleneck imposed by graph-based compression.

\newcolumntype{L}[1]{>{\RaggedRight\arraybackslash}m{#1}}
\newcolumntype{P}[1]{>{\RaggedRight\arraybackslash}p{#1}}

\begin{table*}[t]
\centering
\small
\resizebox{\linewidth}{!}{
\begin{tabular}{@{}L{2.5cm}P{13.5cm}@{}}  
\toprule
\textbf{Field} & \textbf{Content} \\ 
\midrule

\textbf{Query} & 
\emph{``What is the smallest document ID among documents related to an aviation quality control inspector with in-process inspection expertise, a construction manager experienced in oil and gas storage tanks, or collaboration topics?''} \\[0.5em]

\textbf{Expected Answer} & 21 \\[0.5em]
\midrule

\multirow{5}{*}[-2em]{\textbf{HyperGraphRAG}} &
\textbf{Step 1. Search:} “aviation quality control inspector...” → 10 snippets, coherence scores: (1.0, 1.0, 0.5, ..., 0.2) \\[0.4em]
\textbf{} & 
\textbf{Step 2. Search:} “construction manager with oil and gas storage tanks” → 10 snippets, coherence scores: (1.0, 1.0, 1.0, ..., 0.167) \\[0.4em]
\textbf{} &
\textbf{Step 3. Search:} “collaboration topics” → 9 snippets, coherence scores: (1.0, 1.0, 0.5, ..., 0.2) \\[0.4em]
\textbf{} &
\textbf{Step 4. Reasoning:} Since document IDs were missing, the model incorrectly assumed that the most coherent result corresponded to the smallest document ID (ID = 1). \\[0.4em]
\textbf{} &
\textbf{Step 5. Final Answer:} \textcolor{red}{Document ID = 1 (\ding{55})} \\[0.5em]
\midrule

\multirow{5}{*}[-2.5em]{\textbf{MARAG-R1}} &
\textbf{Step 1. Search:} “Aviation quality control inspector...” → [1780, 1793, 1770, 1173] \\[0.4em]
\textbf{} &
\textbf{Step 2. Search:} “Construction manager with oil and gas storage tanks” → [1992, 2027, ..., 1980] (6 docs) \\[0.4em]
\textbf{} &
\textbf{Step 3. Union(Step 1, Step 2)} → [1793, 1988, 1992, ..., 1980] (10 docs) \\[0.4em]
\textbf{} &
\textbf{Step 4. KeywordSearch:} “Collaboration topics” → [21, 37, 154, ..., 1965] (34 docs) \\[0.4em]
\textbf{} &
\textbf{Step 5. Union(Step 3, Step 4)} → 44 unique documents \\[0.4em]
\textbf{} &
\textbf{Step 6. Min(44 documents)} → Document ID = 21 \\[0.4em]
\textbf{} &
\textbf{Step 7. Final Answer:} \textcolor{green}{Document ID = 21 (\ding{51})} \\[0.5em]

\bottomrule
\end{tabular}
}
\caption{Case study of HyperGraphRAG’s failure on OR-query.}
\label{tab:hypergragrag_case_study}
\end{table*}

\subsection{Correlation Between Reward Optimization and Evaluation Performance}

\begin{figure}[t]
\centering
\begin{subfigure}[b]{0.23\textwidth}
\centering
\includegraphics[width=\textwidth]{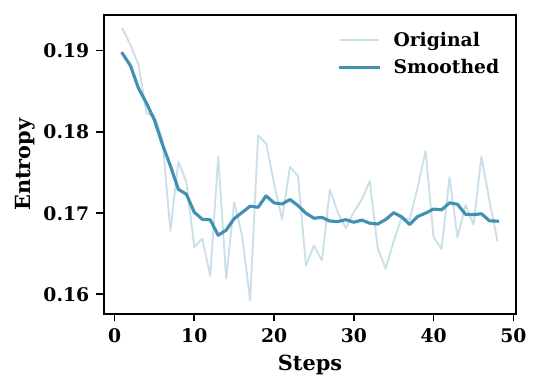}
\caption{Training Entropy}
\label{fig:train_ppl}
\end{subfigure}
\hfill
\begin{subfigure}[b]{0.23\textwidth}
\centering
\includegraphics[width=\textwidth]{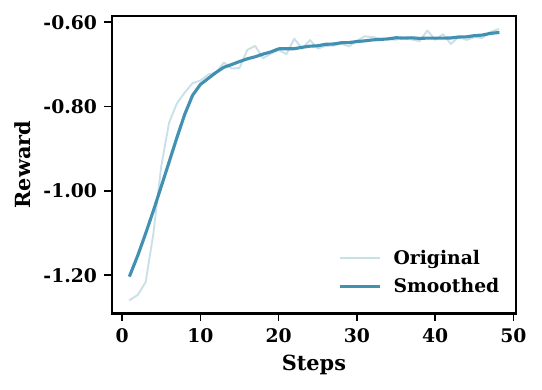}
\caption{Training Reward}
\label{fig:train_reward}
\end{subfigure}
\vspace{0.3cm}
\begin{subfigure}[b]{0.23\textwidth}
\centering
\includegraphics[width=\textwidth]{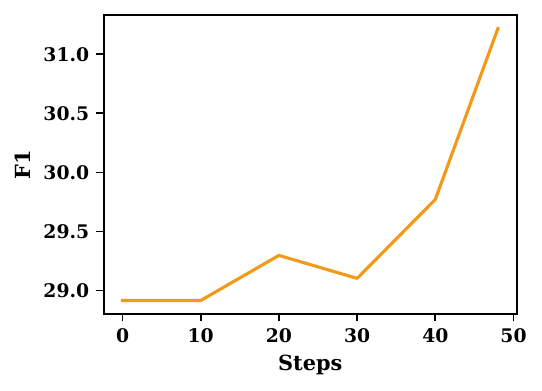}
\caption{Test F1 Score}
\label{fig:test_f1}
\end{subfigure}
\hfill
\begin{subfigure}[b]{0.23\textwidth}
\centering
\includegraphics[width=\textwidth]{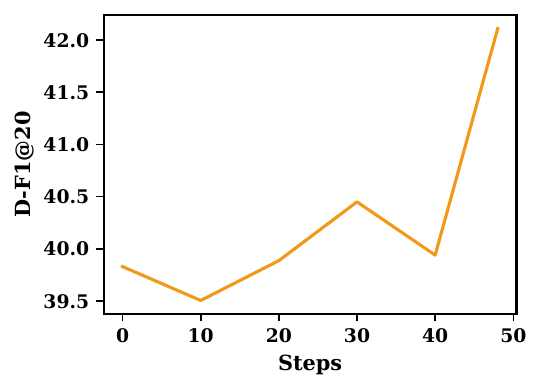}
\caption{Test D-F1@20 Score}
\label{fig:test_ef1}
\end{subfigure}
\caption{Training dynamics during RL optimization. Subfigures show (a) steady decrease in training perplexity indicating improved policy coherence, (b) rapid initial reward growth stabilizing after 40 steps as the policy converges, and (c-d) consistent improvements in both test F1 and D-F1@20 scores validating effective generalization.}
\label{fig:rl_training}
\end{figure}

To verify that our reward design truly drives task-level improvements, we analyze how training rewards evolve alongside evaluation metrics during reinforcement learning. Figure~\ref{fig:rl_training} plots four indicators—training perplexity, training reward, test F1, and test D-F1@20—over training iterations. We observe a clear positive correlation between reward growth and the improvements in test F1 and D-F1@20. As the policy receives higher rewards, it also achieves better evaluation scores, demonstrating that the learned optimization signal aligns well with real task objectives. This confirms that our reward formulation captures the essential aspects of reasoning quality rather than overfitting to superficial behaviors. Furthermore, both F1 and D-F1@20 continue to rise even as rewards plateau, suggesting that the learned policy generalizes effectively beyond the immediate reward signal. Overall, these results show that our two-stage RL framework builds a stable connection between process-level optimization and task-level performance—reward improvement consistently translates into better retrieval completeness and answer accuracy.

\section{Conclusion}
We propose MARAG-R1 to overcome the information-access bottleneck in retrieval-augmented generation. Instead of depending on a single static retriever, MARAG-R1 enables large language models to actively coordinate multiple retrieval tools and iteratively gather external knowledge until the evidence is sufficient for reliable reasoning.
By combining supervised fine-tuning and reinforcement learning, the model learns effective strategies for comprehensive and efficient information acquisition. Experimental results demonstrate that MARAG-R1 substantially improves both retrieval completeness and reasoning accuracy, establishing new state-of-the-art performance on global retrieval tasks. In this way, MARAG-R1 effectively bridges the gap between retrieval and reasoning, unlocking the LLM’s full potential for global understanding.

\appendix

\section{Effect of Different Retrieval Steps}
\begin{figure}[t]
    \centering
    \includegraphics[width=1\linewidth]{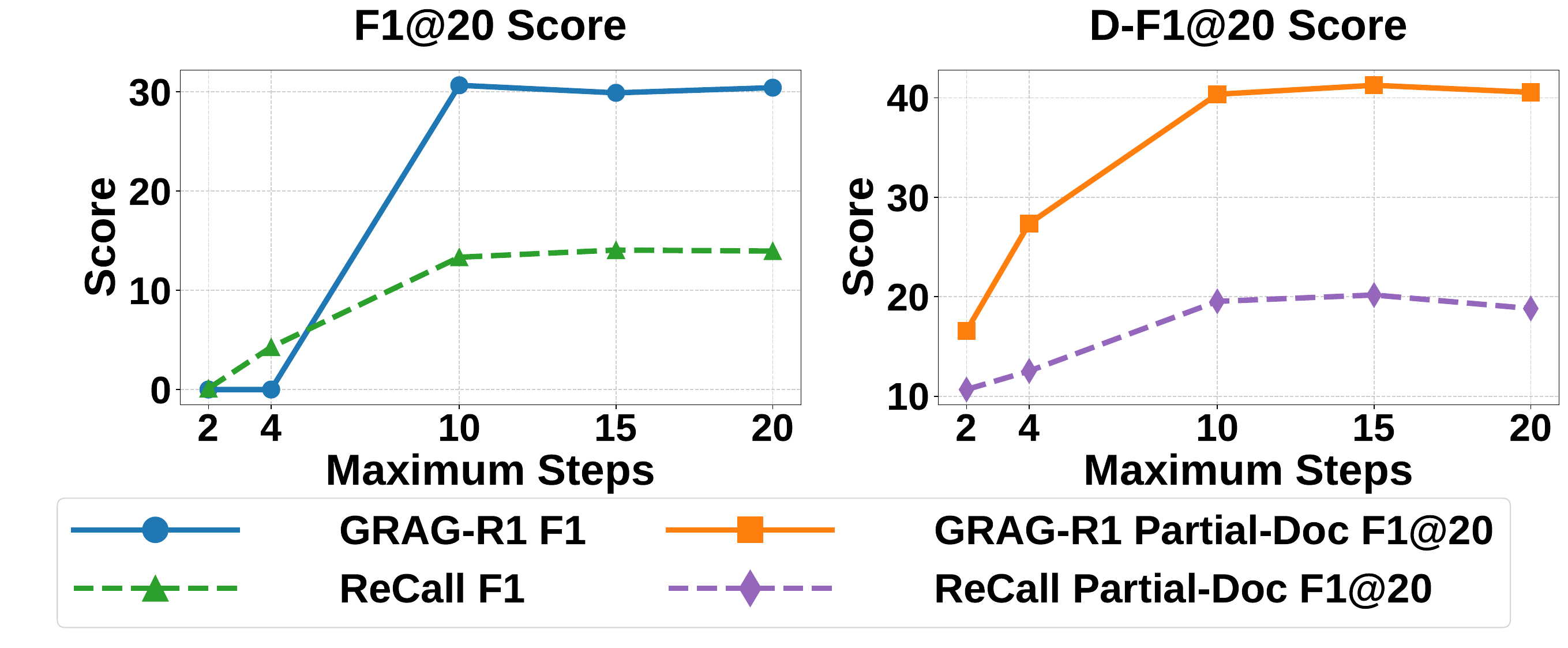}
    \caption{F1/D-F1@20 performance of MARAG-R1 and ReCall under different retrieval steps.}
    \label{fig:different-steps}
\end{figure}

Multi-step retrieval is a fundamental mechanism of our MARAG-R1 framework, enabling iterative evidence gathering and reasoning refinement. As shown in Figure~\ref{fig:different-steps}, which plots F1 and D-F1@20 trends under different maximum retrieval steps, the number of steps directly influences the model’s ability to construct a coherent global knowledge graph. When the step count is small (e.g., 2–5), both MARAG-R1 and ReCall exhibit rapid gains, indicating that iterative retrieval effectively enriches contextual understanding. MARAG-R1’s F1 increases from 0 to 30.65 and D-F1@20 from 16.59 to 40.35, while ReCall shows slower growth, reaching 13.33 and 19.54 respectively. Beyond 10 steps, MARAG-R1 maintains stable performance, suggesting that its reinforcement learning mechanism helps the model balance retrieval depth with noise control. These findings verify that our framework can efficiently leverage additional retrieval steps to achieve more comprehensive reasoning while remaining robust to over-retrieval noise.

\section{Effect of Different Filters}
The retriever defines the quality and coverage of the evidence pool that supports downstream reasoning.
Table~\ref{tab:qwen3-retriever} summarizes the performance of MARAG-R1 when paired with Qwen3 retrievers of different scales.
As the retriever expands from 0.6B to 8B parameters, both F1 and D-F1@20 steadily improve from 31.22/42.11 to 32.31/43.06.
Even the smallest 0.6B retriever performs competitively, indicating that MARAG-R1 can operate effectively with lightweight retrieval backbones.
Meanwhile, larger retrievers provide incremental but consistent gains, suggesting improved evidence coverage and retrieval precision.
These findings confirm that MARAG-R1 generalizes well across retrievers of varying capacities and can flexibly leverage stronger retrievers to enhance global reasoning fidelity.

\begin{table}[ht]
\centering
\renewcommand{\arraystretch}{1.2}
\setlength{\tabcolsep}{8pt}
\begin{tabular}{lcc}
\hline
\textbf{Filter} & \textbf{F1} & \textbf{D-F1@20} \\
\hline
Qwen3-4B  & 31.22 & 42.11 \\
Qwen3-8B  & 31.58 & 42.85 \\
Qwen3-14B & 31.82 & 43.16 \\
\hline
\end{tabular}
\caption{Performance of MARAG-R1 under different Qwen3 filter sizes.}
\label{tab:qwen3-filter}
\end{table}

\section{Effect of Different Retrievers}
The retriever determines the quality and completeness of the evidence pool from which reasoning is built. Table~\ref{tab:qwen3-retriever} reports the performance of MARAG-R1 with Qwen3 retrievers of different parameter scales. As the retriever grows from 0.6B to 8B, F1 and D-F1@20 steadily improve from 29.73/40.17 to 30.57/41.26. The small 0.6B retriever already performs competitively, confirming that MARAG-R1 can operate effectively even with lightweight retrieval backbones. Meanwhile, larger retrievers provide marginal yet consistent gains, suggesting enhanced coverage and retrieval precision. Overall, these results indicate that our method generalizes well across retrievers of various capacities and can flexibly leverage stronger retrievers to further enhance global reasoning fidelity.

\begin{table}[ht]
\centering
\renewcommand{\arraystretch}{1.2}
\setlength{\tabcolsep}{8pt}
\begin{tabular}{lcc}
\hline
\textbf{Retriever} & \textbf{F1} & \textbf{D-F1@20} \\
\hline
Qwen3-0.6B & 31.22 & 42.11 \ \\
Qwen3-4B   & 31.45 & 42.12 \\
Qwen3-8B   & 32.31 & 43.06 \\
\hline
\end{tabular}
\caption{Performance of MARAG-R1 under different Qwen3 retriever sizes.}
\label{tab:qwen3-retriever}
\end{table}

\bibliography{ref}

\end{document}